\documentclass[runningheads]{llncs}

\usepackage{eccv}

\usepackage{eccvabbrv}
\usepackage{makecell}
\usepackage{graphicx}
\usepackage{booktabs}
\usepackage{pifont}
\usepackage{bbding}
\usepackage{graphicx}
\usepackage{wrapfig}
\usepackage{caption}
\usepackage{float}
\usepackage[accsupp]{axessibility}  

\usepackage[colorlinks]{hyperref}

\usepackage{orcidlink}

\begin{document}

\title{Progressive Pretext Task Learning for\\ Human Trajectory Prediction} 


\author{Xiaotong Lin\inst{1} \and
Tianming Liang\inst{1} \and
Jianhuang Lai\inst{1,2,3} \and
Jian-Fang Hu\inst{1,2,3}*}

\authorrunning{X.~Lin et al.}

\institute{School of Computer Science and Engineering, Sun Yat-sen University, China \and
Guangdong Province Key Laboratory of Information Security Technology, China \and
Key Laboratory of Machine Intelligence and Advanced Computing, Ministry of Education, China \\
\email{\{linxt29, liangtm\}@mail2.sysu.edu.cn, \{stsljh, hujf5\}@mail.sysu.edu.cn}
}

\maketitle
\let\thefootnote\relax\footnotetext{* Corresponding Author.}

\begin{abstract}
Human trajectory prediction is a practical task of predicting the future positions of pedestrians on the road, which typically covers all temporal ranges from short-term to long-term within a trajectory. However, existing works attempt to address the entire trajectory prediction with a singular, uniform training paradigm, neglecting the distinction between short-term and long-term dynamics in human trajectories. To overcome this limitation, we introduce a novel Progressive Pretext Task learning (PPT) framework, which progressively enhances the model's capacity of capturing short-term dynamics and long-term dependencies for the final entire trajectory prediction. Specifically, we elaborately design three stages of training tasks in the PPT framework. In the first stage, the model learns to comprehend the short-term dynamics through a stepwise next-position prediction task. 
In the second stage, the model is further enhanced to understand long-term dependencies through a destination prediction task. 
In the final stage, the model aims to address the entire future trajectory task by taking full advantage of the knowledge from previous stages. To alleviate the knowledge forgetting, we further apply a cross-task knowledge distillation. Additionally, we design a Transformer-based trajectory predictor, which is able to achieve highly efficient two-step reasoning by integrating a destination-driven prediction strategy and a group of learnable prompt embeddings. Extensive experiments on popular benchmarks have demonstrated that our proposed approach achieves state-of-the-art performance with high efficiency. Code is available at \href{https://github.com/iSEE-Laboratory/PPT}{https://github.com/iSEE-Laboratory/PPT}.
\end{abstract}

\section{Introduction}
\label{sec:intro}

Human trajectory prediction has found extensive applications in various critical domains, such as autonomous driving~\cite{levinson2011towards,choi2021shared,park2020diverse,salzmann2020trajectron++}, surveillance systems~\cite{valera2005intelligent}, robotic navigation~\cite{foka2010probabilistic,li2020end} and planning~\cite{luo2018porca,song2020pip}. Given an observed human trajectory, the objective of human trajectory prediction is to precisely forecast the unobserved plausible future trajectories. 
This includes predicting positions from short-term to long-term future, covering all temporal ranges within a trajectory.

\begin{figure}[!t]
\centering
\includegraphics[width=\linewidth]{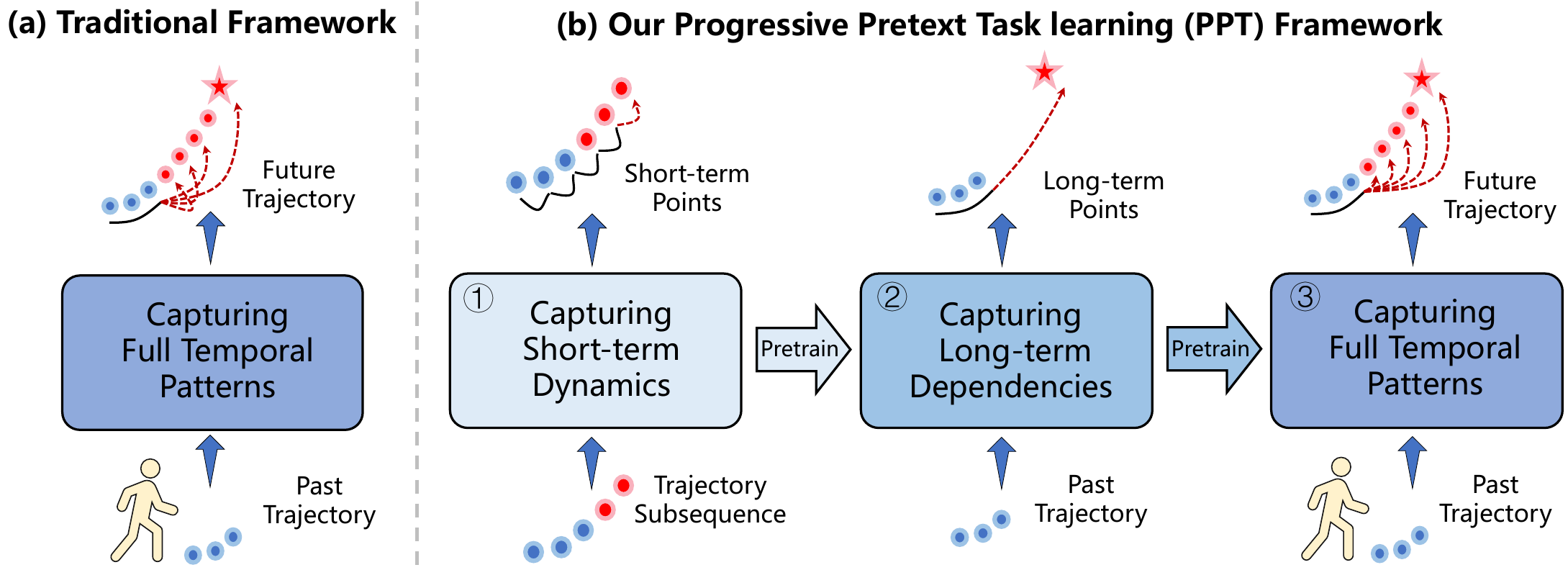}
\caption{Comparing our Progressive Pretext Task learning (PPT) framework with regular trajectory prediction. (a) Existing works tend to aggressively force the model to learn complicated full temporal patterns at once. (b) Our Framework employs three learning stages to progressively enhance the model for future trajectory prediction.}
\label{fig:intro_comp}
\end{figure}

Generally, making predictions across different temporal distances relies on distinct aspects of understanding abilities.
On one hand, short-term future prediction requires to recognize the local dynamic patterns from the immediate, fine-grained variations between timesteps.
On the other hand, long-term future prediction aims to deduce the global tendencies by capturing long-term dependencies of trajectories.
However, this distinction is always neglected by existing methods~\cite{gupta2018social,tsao2022social,mohamed2020social,shi2021sgcn,xu2022socialvae,mao2023leapfrog,gu2022stochastic}.
They attempt to address both short-term and long-term prediction with a singular, uniform training paradigm, often struggling in a suboptimal compromise between short-term and long-term performance.

To overcome this limitation, in this paper, we present a novel Progressive Pretext Task learning (PPT) framework, which progressively enables the model to capture the complicated short-term dynamics and long-term dependencies for the entire future trajectory prediction. To be specific, PPT comprises three stages of progressive training tasks. \underline{Task-\uppercase\expandafter{\romannumeral1}} aims to equip the model with the basic capacity to comprehend short-term dynamics inherent in the trajectories, by predicting the next position given a trajectory of arbitrary length. \underline{Task-\uppercase\expandafter{\romannumeral2}} intends to enhance the model to capture long-term dependencies by predicting the destinations of trajectories, where a diversity loss is employed to encourage intention diversity of a pedestrian. Once pretext Task-\uppercase\expandafter{\romannumeral1} and Task-\uppercase\expandafter{\romannumeral2} are completed, the model acquires the ability to capture both short-term dynamics and long-term dependencies within a trajectory. Given this, in \underline{Task-\uppercase\expandafter{\romannumeral3}}, we take full advantage of the enhanced knowledge for more accurate prediction, by finetuning the well-pretrained model from Task-\uppercase\expandafter{\romannumeral2} for entire future trajectory prediction. Moreover, to preserve the knowledge acquired from previous pretext tasks and stabilize prediction performance, we introduce cross-stage knowledge distillation, transferring the knowledge of Task-\uppercase\expandafter{\romannumeral1} and Task-\uppercase\expandafter{\romannumeral2} into the models in Task-\uppercase\expandafter{\romannumeral3}.

In our PPT framework, we further devise a novel Transformer-based trajectory predictor. Compared to previous Transformer predictors \cite{sadeghian2019sophie,yu2020spatio,yuan2021agentformer,gu2022stochastic,giuliari2021transformer} that autoregressively generate the future positions, our model is able to efficiently predict the trajectory of any length in only two steps: determining the destination firstly and then generating the rest future points all at once. Specifically, our model consists of a destination predictor and a trajectory predictor. The former predictor aims to capture long-term dependencies for predicting destinations, which are used to guide the latter one in generating the entire future trajectories. 
To achieve the efficient parallel generation of trajectory points, we introduce a series of learnable prompt embeddings to indicate the certain timesteps.

Extensive experiments demonstrate that our framework achieves state-of-the-art results on various popular datasets, validating the superiority of our framework. Moreover, ablation studies are conducted to verify the effectiveness of each pretext task and other key components. Qualitatively, our framework can produce human trajectories that are more accurate and temporally acceptable. 

Overall, our contributions are summarized as follows:
\begin{itemize}
\item We present PPT, a novel progressive pretext task learning framework to progressively enable the model to capture the complicated dependencies across various temporal ranges in human trajectories, including short-term dynamics and long-term dependencies, for the entire future trajectory prediction.

\item We propose a Transformer-based trajectory predictor, which adopts a two-step destination-driven strategy and integrates a series of learnable prompts to achieve effective and efficient prediction.

\item Extensive experiments on four commonly used datasets demonstrate that our framework can consistently outperform the current state-of-the-art methods.
\end{itemize}

\section{Related Work}

Human trajectory prediction aims to forecast the reasonable future path given an observed sequence of movements. Considering the indeterminate nature of human motion, this task is particularly challenging due to the necessity of predicting the precise coordinates of the positions over all timesteps, which requires addressing both the short-term dynamics and long-term dependencies. 

\subsection{Human Trajectory Prediction}

Existing works can be briefly divided into two branches: one branch focuses on the utilization of scene maps~\cite{mangalam2021goals,lee2022muse,yue2022human,sadeghian2019sophie,wong2022view}, while the other aims to mine the movement patterns and interactions~\cite{gupta2018social,xu2022socialvae,mohamed2020social,tsao2022social,kosaraju2019social,shi2021sgcn,mangalam2020not,xu2022remember,sun2021three,wong2023msn,xie2024pedestrian}. Considering the computational costs of modeling scene maps, in this paper, we follow the latter branch to explore a more effective approach for understanding temporal movement patterns within trajectories.
To address this task, a lot of efforts have been made. For example, Gupta et al. \cite{gupta2018social} initially proposed to utilize a GAN-based \cite{goodfellow2014generative} network, and train the model by directly aligning various temporal positions in future trajectories with GT without differentiation. Gu et al. \cite{gu2022stochastic} employ a Transformer-based diffusion network and train the model to produce the entire future trajectory at once. However, these works overlook the differences between the learning patterns of short-term and long-term prediction, which cause suboptimal performance during joint optimization. 
While recent destination-based methods \cite{mangalam2020not,xu2022remember,zhao2021you} attempt to alleviate this issue by initially predicting the destination with one predictor and then interpolating intermediate positions with another, they overlook the knowledge transfer between destination prediction and intermediate position prediction, which results in a significant gap between the destination predictor and the trajectory predictor. To overcome the limitations, in our paper, we devise a Progressive Pretext Task learning framework, which introduces two well-designed pretext tasks to incrementally enhance the model to capture both short-term dynamics and long-term dependencies for the entire future trajectory prediction. 

\subsection{Transformer-based Human Trajectory Prediction}

In recent years, Transformer \cite{vaswani2017attention,tang2023predicting} architectures have demonstrated impressive capability in capturing complex sequential dependencies. Considering its effectiveness, researchers \cite{sadeghian2019sophie,yu2020spatio,yuan2021agentformer,gu2022stochastic,giuliari2021transformer,shi2023trajectory} have increasingly turned to Transformer for human trajectory prediction. For example, STAR \cite{yu2020spatio} modeled the crowd as a graph and leveraged a graph-based Transformer to learn the spatiotemporal interaction of the crowd motion.
Also, Tsao et al. \cite{tsao2022social} use a Transformer as a backbone model and propose some pretext tasks regarding cross-sequence modeling. However, they are always inefficient during inference since they generate the trajectory points in an autoregressive manner. Recently, MID and TUTR have attempted to explore the non-autoregressive Transformer in this task. Nevertheless, MID \cite{gu2022stochastic} relies on a diffusion model, which significantly increases the inference time. TUTR \cite{shi2023trajectory} ignores the temporal motion dynamics in the trajectory, leading to suboptimal performance.
In this work, we propose a novel non-autoregressive Transformer to overcome the above limitations. 
Compared to TUTR, our model introduces a series of effective learnable prompts to represent unobserved positions, which significantly improves the prediction performance.

\subsection{Progressive Pretraining}

So far, progressive learning techniques have been explored in a wide range of tasks, including image generation~\cite{gregor2015draw,karras2017progressive}, image enhancement~\cite{fu2023you,liang2023iterative}, object detection~\cite{cai2018cascade,gidaris2015object,komodakis2016attend,najibi2016g} and motion prediction~\cite{tang2024temporal,ma2022progressively}. Specifically, Karras et al. \cite{karras2017progressive} proposed to start with low-resolution images, and then progressively increase the resolution by adding layers to the networks. PGBIG~\cite{ma2022progressively} utilize multiple stages to progressively refine the initial guess of the future frames. Fu et al.~\cite{fu2023you} introduce a progressive learning strategy for low-light image enhancement. In the process of self-knowledge distillation, they gradually increase the proportion of low-light images as input to the student branch, aiming to progressively enhance the learning difficulty for the student. However, progressive pretraining remains unexplored in the community of human trajectory prediction. To the best of our knowledge, our work is the first to explore progressive pretraining in human trajectory prediction, introducing two well-designed pretext tasks to incrementally enable the model to capture short-term dynamics and long-term dependencies for the entire future trajectory prediction.

\begin{figure*}[!t]
\centering
\includegraphics[width=\textwidth]{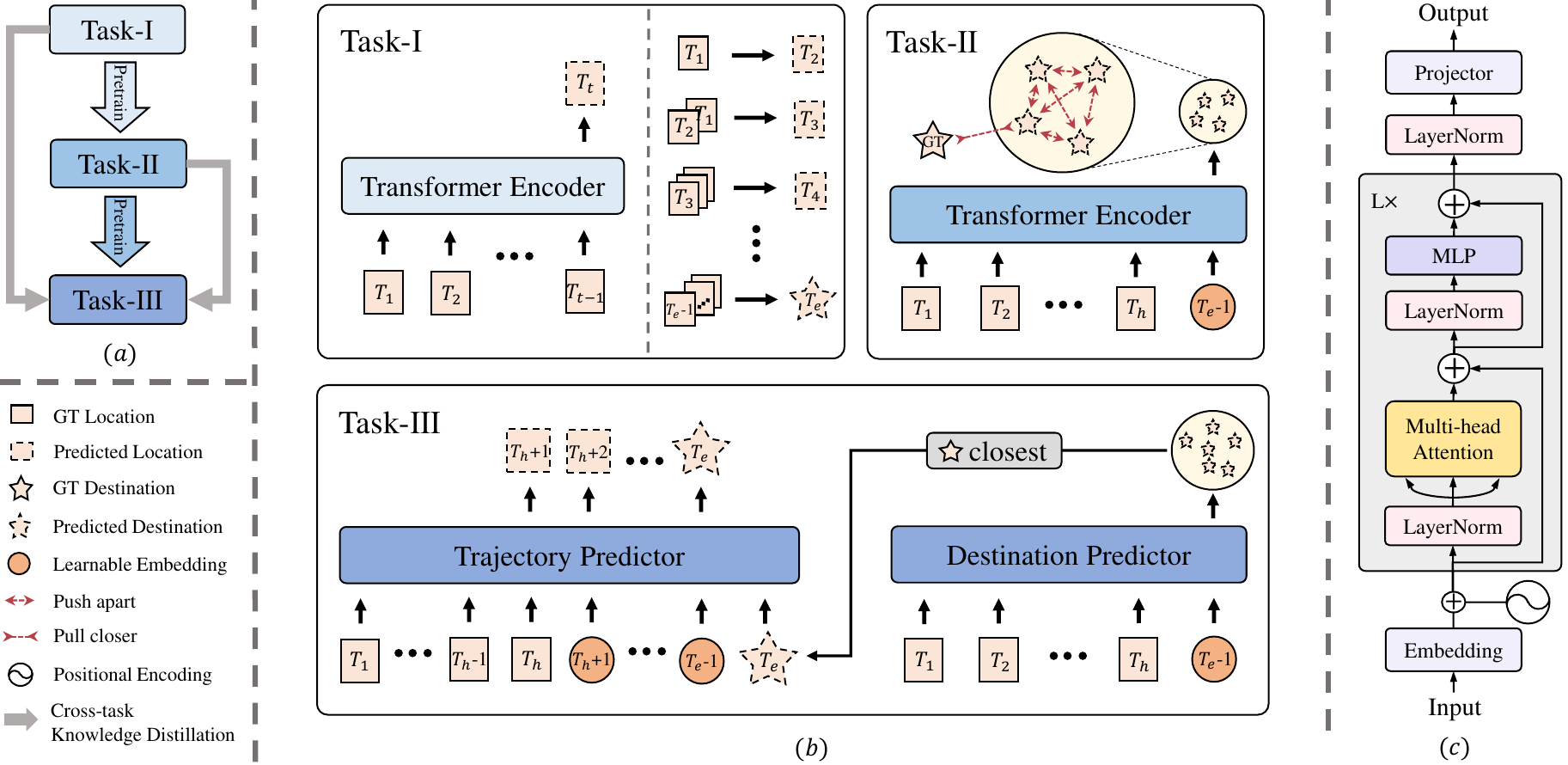}
\caption{Illustration of our overall Progressive Pretext Task learning (PPT) framework. (a) demonstrates our progressive training pipeline, where each training stage employs a corresponding task to incrementally enhance the model's capacity for the entire future trajectory prediction. A cross-task knowledge distillation is introduced to avoid knowledge forgetting. Specifically, as in (b), we sequentially perform the stepwise next-position prediction (Task-\uppercase\expandafter{\romannumeral1}), the leapfrog destination prediction (Task-\uppercase\expandafter{\romannumeral2}) and the complete trajectory prediction (Task-\uppercase\expandafter{\romannumeral3}). (c) shows our backbone model.}
\label{fig:architecture}
\end{figure*}

\section{Method}
\label{sec:method}

\noindent\textbf{Problem Formulation.}
Human trajectory prediction aims to accurately predict the future trajectory based on the observation of past trajectory, with the key challenge of capturing both short-term dynamics and long-term dependencies.
Formally, given a series of past observed trajectories presented as $\mathcal{S}^{T_1:T_h}=\{(x^{T_1},y^{T_1}), ..., (x^{T_h}, y^{T_h})\}_{n=1}^N$ for N agents over time $T_1, T_2,..., T_h$, the target of human trajectory prediction is to forecast the subsequent 2D positions for the unobserved future $\mathcal{S}^{T_h+1:T_h+T_f}=\{(x^{{T_h+1}}, y^{T_h+1}), ...,$ $ (x^{T_h+T_f}, y^{T_h+T_f})\}_{n=1}^N$. In the following, we denote $T_e=T_h+T_f$ as the entire trajectory length.

\noindent\textbf{Overview.} As shown in Figure~\ref{fig:architecture}, we propose a Progressive Pretext Task learning (PPT) framework for trajectory prediction, aiming to incrementally enhance the model's capacity to understand the past trajectory and predict the future trajectory.
Specifically, our framework consists of three stages of progressive training tasks, as illustrated in Figure~\ref{fig:architecture} (b). In Stage I, we pretrain our predictor on pretext Task-\uppercase\expandafter{\romannumeral1}, aiming to fully understand the short-term dynamics of each trajectory, by predicting the next position of a trajectory of arbitrary length. In Stage II, we further train this predictor on pretext Task-\uppercase\expandafter{\romannumeral2}, intending to capture the long-term dependencies, by predicting the destination of a trajectory.
Once Task-\uppercase\expandafter{\romannumeral1} and Task-\uppercase\expandafter{\romannumeral2} are completed, the model is capable of capturing both the short-term dynamics and long-term dependencies within the trajectory. Finally, in Stage III, we duplicate our model to obtain two predictors: one for destination prediction and the other for intermediate waypoint prediction. In this stage, we perform Task-\uppercase\expandafter{\romannumeral3} that enables the model to achieve the complete pedestrian trajectory prediction.
For the sake of stable training, we further employ a cross-task knowledge distillation to avoid knowledge forgetting.

\noindent\textbf{Backbone.}
In this work, we employ a Transformer encoder~\cite{vaswani2017attention} as our backbone model, which is shown in Figure \ref{fig:architecture} (c). Given the 2D positions as input, \eg, trajectory sequences $S^{T_{m}:T_{n}}$ from time $T_m$ to $T_n$, we first use an embedding layer to convert them to input features. Then, these features with corresponding temporal position embeddings \{$T_m,T_{m}+1,...,T_n$\} are passed through multiple Transformer layers, each consisting of pre-norm \cite{DBLP:conf/acl/WangLXZLWC19}, multi-head attention, LayerNorm (LN) and an MLP. 
The model learns to understand the trajectory through the feature interactions between different positions, and outputs the interactive representation for each position.
These outputs are fed into a final LN, followed by a linear projector to obtain the future 2D positions $\hat{S}^{T_{m}+1:T_{n}+1}$, which represents the predicted next-frame positions corresponding to each input location.
Specially for Task-\uppercase\expandafter{\romannumeral2} and Task-\uppercase\expandafter{\romannumeral3}, we employ learnable prompt embeddings to represent the unobserved future positions in the trajectory, as illustrated in~\Cref{fig:architecture} (b). Details will be presented in the following.

\subsection{Task-\uppercase\expandafter{\romannumeral1}: Stepwise Next-position Prediction}
\label{subsec:task1}

Given an observed trajectory sequence of arbitrary length, the target of the first pretext task is to accurately predict the position of the next point. This task promotes the model to explore the motion patterns and understand the short-term dynamics of each pedestrian trajectory.

Specifically, for the trajectory sequence $\mathcal{S}^{T_1:T_e}$, we randomly sample its subsequence $\mathcal{S}^{T_1:T_{t-1}}$, and then feed it into the model $\theta$ to infer the next position $\mathcal{S}^{T_t}$, as shown in Figure~\ref{fig:architecture} (b).
The randomness can bring the effect of data augmentation.
In practice, multiple random subsequences would be sampled from one trajectory for improving training efficiency, and this can be effectively implemented by leveraging the causal self-attention mask \cite{radford2018improving}. We use $\theta_I$ to indicate the model trained with Task-\uppercase\expandafter{\romannumeral1}.

While this task is simple and straightforward, it can effectively enable the model to identify the motion patterns and capture the short-term dynamics within the trajectory. As understanding the patterns and dynamics of the trajectory is an essential capacity for trajectory prediction, this knowledge can be transferred and further exploited to facilitate the prediction in later tasks.

\subsection{Task-\uppercase\expandafter{\romannumeral2}: Leapfrog Destination Prediction}
\label{subsec:task2}
The target of the second pretext task is to predict the destination of a trajectory. This is challenging since it requires the model to speculate the pedestrian moving intention from the past trajectory and capture the long-term dependency between the final destination and the early trajectory.

Specifically, taking the past trajectory sequence $\mathcal{S}^{T_1:T_h}$ as input, Task-\uppercase\expandafter{\romannumeral2} continues to train $\theta_I$ for predicting the destination $\mathcal{S}^{T_e}$ of the entire trajectory. 
Considering the inherent indeterminate nature of human motion, we follow the previous works~\cite{gupta2018social,xu2022remember,xu2022socialvae} to predict multiple destinations (\eg, K) once. In practice, we feed the output feature at the destination to an MLP to regress K destinations. In order to ensure prediction accuracy, we incorporate a precision loss \cite{gupta2018social} to minimize the distance between the ground truth destination $\mathbf{E}$ and its closest predicted destination, which is formed as $ L_{Precision} = \mathop{\min}_{k} L_2(\mathbf{\hat{E}}_k, \mathbf{E})$. Here, $L_2$ is the Euclidean distance function.
Furthermore, to prevent K predicted destinations from falling into the same modality, we employ a diversity loss as \cite{xu2022diverse, yuan2020dlow} to provide sufficient diversity. As shown in Figure \ref{fig:architecture} (b), we promote the pairwise distance between the trajectory destinations as follows:
\begin{align}
    \label{eq:Div}
    \begin{split}
        L_{Diversity} = \frac{1}{K(K-1)} \sum^K_i\sum^K_{j\neq i} e^{-\frac{L_2^{^2}(\mathbf{\hat{E}}_i, \mathbf{\hat{E}}_j)}{\sigma_s}},
    \end{split}
\end{align}
where $\sigma_s$ is a scaling factor. With this diversity loss, the model can produce more diverse destinations, thus leading to more diverse trajectories. 

The loss function for destination prediction in this task is demonstrated as:
\begin{equation}
\label{eq:Des}
    L_{Des} = L_{Precision} + \lambda_d L_{Diversity},
\end{equation}
where $\lambda_d$ balances the accuracy and diversity of the predicted destinations.

Notably, to align with the input of model $\theta_{\uppercase\expandafter{\romannumeral1}}$, we assign a corresponding positional encoding to each position. However, due to the absence of the ground truth data for future trajectory, the ($T_e-1$)-th position cannot be accessed as input to predict the $T_e$-th position (destination). Therefore, we introduce a learnable prompt embedding and append it after the past trajectory sequence, aiming to predict destinations in a leapfrog manner. We further set the positional encoding for this learnable embedding as $T_e-1$ to maintain consistency with Task-\uppercase\expandafter{\romannumeral1}, indicating its prediction for the $T_e$-th position (destination), as in~\Cref{fig:architecture} (b).

Through this leapfrog destination prediction task, the well-trained model $\theta_{\uppercase\expandafter{\romannumeral2}}$ can acquire the ability for long-term prediction, which can provide the guiding reference and the knowledge associated with the long-term dependencies for the entire future trajectory prediction.

\subsection{Task-\uppercase\expandafter{\romannumeral3}: Comprehensive Trajectory Prediction}
\label{subsec:task3}

With the training on Task-\uppercase\expandafter{\romannumeral1} and Task-\uppercase\expandafter{\romannumeral2}, the model $\theta_{II}$ has the capacity of understanding short-term dynamics (acquired from Task-\uppercase\expandafter{\romannumeral1}) and capturing long-term dependencies within the future trajectory (acquired from Task-\uppercase\expandafter{\romannumeral2}). In the final task, we take full advantage of this knowledge for the complete trajectory prediction task: predicting all the positions within the future trajectories.

To be specific, we replicate the model $\theta_{II}$ into a destination predictor and a trajectory predictor, as illustrated in Figure \ref{fig:architecture} (b). 
We employ the destination predictor to generate K candidate destinations, as mentioned in Task-\uppercase\expandafter{\romannumeral2}, and then feed the one closest to the ground truth (GT) into the trajectory predictor. 
The input sequence of the trajectory predictor can be divided into three parts: the observed trajectory from $T_1$ to $T_h$, the unobserved future trajectory from $T_{h}+1$ to $T_{e}-1$, and the pseudo destination at $T_{e}$. Specially for the unobserved future trajectory, we use learnable prompt embeddings as input. With these inputs, the trajectory predictor outputs the 2D positions for the entire future trajectory, \ie, $\mathcal{S}^{T_h+1:T_e}$. 
During Task-\uppercase\expandafter{\romannumeral3}, we jointly train the destination predictor and trajectory predictor to regress the entire future trajectory.

To avoid the knowledge from previous pretext tasks being forgotten, we devise a cross-task knowledge distillation for additional regularization in Task-\uppercase\expandafter{\romannumeral3}. Specifically, we punish the output differences between $\theta_{\uppercase\expandafter{\romannumeral1}}$ and the trajectory predictor, as well as $\theta_{\uppercase\expandafter{\romannumeral2}}$ and the destination predictor, respectively, with the following loss functions:
\begin{align}
    \begin{split}
        L^t_{kd} &= ||\mathbf{\mathcal{F}}_{\uppercase\expandafter{\romannumeral1}}^t - \mathcal{P}_t(\mathbf{\mathcal{F}}^t_{\uppercase\expandafter{\romannumeral3}})||_2,      \\
        L^d_{kd} &= ||\mathbf{\mathcal{F}}_{\uppercase\expandafter{\romannumeral2}}^d 
- \mathcal{P}_d(\mathbf{\mathcal{F}}_{\uppercase\expandafter{\romannumeral3}}^d)||_2,
    \end{split}
\end{align}
where $\mathbf{\mathcal{F}}_i^t$ and $\mathbf{\mathcal{F}}_i^d$ indicates the output features of future trajectory and destination obtained in $i$-th task, respectively. $\mathcal{P}_t$ and $\mathcal{P}_d$ denotes the linear projector.

Overall, the loss function in this stage is formulated as:
\begin{equation}
\label{eq:Traj}
    L_{Traj} = L_{Recon} + \lambda^{t}_{kd} L^t_{kd} + \lambda^{d}_{kd} L^d_{kd},
\end{equation}
where $L_{Recon}$ is the $L_2$ distance between the predicted and ground truth future trajectory. The $\lambda^t_{kd}$ and $\lambda^d_{kd}$ are leveraged to control the trade-off between different loss terms.

\subsection{Inference}

After training on all three tasks that progressively enable the model to predict the entire future trajectory, we employ the well-trained destination predictor and trajectory predictor in the final stage for inference. Specifically, we first utilize the destination predictor to predict K destinations. Then, we take each of these destinations as the input to the trajectory predictor, guiding the generation of K future trajectories.

\section{Experiments}
\label{sec:exp}

In this section, we conduct extensive experiments on various popular pedestrian trajectory prediction benchmark datasets. The results show that our approach consistently outperforms the current state-of-the-art methods quantitatively and qualitatively. Further, ablation studies are provided to demonstrate the effectiveness of the key components in our proposed framework.

\begin{table*}[!t] 
\renewcommand{\arraystretch}{1.1}
    \centering
    \caption{Comparisons with the current state-of-the-art methods on the SDD dataset in minADE$_{20}$ / minFDE$_{20}$ (pixels) metric. Text in \textbf{bold} denotes the best results. Our method outperforms other approaches by a large margin.}
    \resizebox{\textwidth}{!}{
    \begin{tabular}{c|c|c|c|c|c|c|c|c|c|c}
        \toprule[2pt]
        Method & \makecell[c]{Social\\-GAN \cite{gupta2018social}} & \makecell[c]{SOPHIE\\ \cite{sadeghian2019sophie}} & \makecell[c]{PECNet\\ \cite{mangalam2020not}} & \makecell[c]{PCCSNet\\ \cite{sun2021three}} & \makecell[c]{MemoNet\\ \cite{xu2022remember}} & \makecell[c]{Social-\\VAE \cite{xu2022socialvae}} & MID \cite{gu2022stochastic} & LED \cite{mao2023leapfrog} & TUTR \cite{shi2023trajectory} & PPT (Ours) \\
        \hline
        ADE ↓  & 27.23  & 16.27   & 9.96    & 8.62   & 8.56   & 8.10    & 7.61   & 8.48   & 7.76    & \textbf{7.03}             \\
        FDE ↓  & 41.44  & 29.38   & 15.88   & 16.16  & 12.66  & 11.72  & 14.30   & 11.66  & 12.69   & \textbf{10.65}             \\
        \bottomrule[2pt]
    \end{tabular}}
    \label{tab:SDD}
\end{table*}

\subsection{Experimental Setup}
\textbf{Datasets.} Our proposed PPT framework is evaluated on four widely used public pedestrian datasets: Stanford Drone Dataset (SDD) \cite{robicquet2016learning}, ETH \cite{pellegrini2009you}/UCY \cite{leal2014learning} dataset and Grand Central Station (GCS) \cite{yi2015understanding} dataset. SDD is one of the most popular benchmarks which is a large-scale dataset recorded by drone cameras in bird's eye view. It contains trajectories of 5,232 pedestrians in eight different scenes. The ETH/UCY is a combination of two datasets with five different scenes. The ETH~\cite{pellegrini2009you} dataset contains two scenes, ETH and HOTEL, with 750 pedestrians, and the UCY~\cite{leal2014learning} is composed of three scenes with 786 pedestrians, including UNIV, ZARA1 and ZARA2. The GCS dataset captures a complex and densely populated scene within one of the largest and busiest train stations in the United States. This dataset includes trajectories of 12,684 pedestrians over a duration of approximately one hour.

\noindent\textbf{Evaluation Metrics.}
We employ the same data processing procedure and evaluation configuration as the previous works \cite{xu2022remember,gupta2018social,mangalam2020not,bae2022non}. For performance evaluation, we adopt the Average Displacement Error (ADE) and Final Displacement Error (FDE) as evaluation metrics, which measure the average position distance and the destination distance between the predicted trajectories and the ground truth (GT) trajectories, respectively. Considering the inherent uncertainty of the future and the indeterminate nature of human motion, we generate K=20 future trajectories for every past trajectory and calculate the minimum ADE and FDE (Best-of-20 strategy) performance as in the prior works \cite{gupta2018social,gu2022stochastic,mangalam2020not,xu2022remember,bae2022non}. For all datasets, we take the past 8 steps (3.2s) as the observed trajectory and predict the following future 12 steps (4.8s).

\noindent\textbf{Implementing Details.} In our implementation, the Transformer encoder in all stages comprises three layers, where the Transformer dimension is set to 128, and 8 attention heads are applied. The scaling factor $\sigma_s$ in Equation (\ref{eq:Div}) is assigned a value of 1, and the weight hyperparameter $\lambda_d$ in Equation (\ref{eq:Des}) is set to 100. We let $\lambda^t_{kd}=5$ and $\lambda^d_{kd}=0.5$ in Equation (\ref{eq:Traj}). To retain the knowledge acquired from Task-\uppercase\expandafter{\romannumeral1} to the fullest extent, in training Stage-\uppercase\expandafter{\romannumeral2}, we initially train a Multi-Layer Perceptron (MLP) for destination regression as a warm-up, and then jointly train the entire model. We employ the Adam optimizer \cite{kingma2014adam} for all three training stages, with the learning rate set to \{0.001, 0.0001, 0.0015\} respectively. All of our experiments were conducted using PyTorch on a single RTX 3090 GPU.

\begin{table*}[!ht] 
\renewcommand{\arraystretch}{1.1}
    \centering
    \caption{Comparisons with the current state-of-the-art methods on the ETH/UCY dataset in minADE$_{20}$ / minFDE$_{20}$ (meters) metric. Text in \textbf{bold} denotes the best results. Among all the methods, our proposed approach achieves the best performance.}
    \resizebox{\textwidth}{!}{
    \begin{tabular}{c||c|c||c|c|c||c}
        \toprule[2pt]
        Method & ETH & HOTEL & UNIV & ZARA1 & ZARA2 & AVG \\
        \hline
        Social-GAN \cite{gupta2018social}   & 0.87/1.62   & 0.67/1.37  & 0.76/1.52  & 0.35/0.68      & 0.42/0.84    & 0.61/1.21             \\
        STAR \cite{yu2020spatio}            & 0.36/0.65   & 0.17/0.36  & 0.31/0.62  & 0.29/0.52      & 0.22/0.46    & 0.26/0.53             \\
        PECNet \cite{mangalam2020not}       & 0.54/0.87   & 0.18/0.24  & 0.35/0.60  & 0.22/0.39      & 0.17/0.30    & 0.29/0.48             \\
        AgentFormer \cite{yuan2021agentformer}   & 0.45/0.75   & 0.14/0.22  & 0.25/0.45  & 0.18/0.30      & 0.14/0.24    & 0.23/0.39              \\
        PCCSNet \cite{sun2021three}         & \textbf{0.28}/0.54   & \textbf{0.11}/0.19  & 0.29/0.60  & 0.21/0.44      & 0.15/0.34    & 0.21/0.42              \\
        MemoNet \cite{xu2022remember}       & 0.40/0.61   & \textbf{0.11}/0.17  & 0.24/0.43  & 0.18/0.32      & 0.14/0.24    & 0.21/0.35              \\
        MID \cite{gu2022stochastic}         & 0.39/0.66   & 0.13/0.22  & 0.22/0.45  & \textbf{0.17}/0.30      & 0.13/0.27    & 0.21/0.38              \\
        SocialVAE \cite{xu2022socialvae}    & 0.41/0.58   & 0.13/0.19  & \textbf{0.21}/\textbf{0.36}  & \textbf{0.17}/0.29     & 0.13/0.22     & 0.21/0.33   \\
        LED \cite{mao2023leapfrog}          & 0.39/0.58   & \textbf{0.11}/0.17  & 0.26/0.43  & 0.18/\textbf{0.26}     & 0.13/0.22     & 0.21/0.33   \\
        NPSN \cite{bae2022non}              & 0.36/0.59      & 0.16/0.25    & 0.23/0.39      & 0.18/0.32       & 0.14/0.25    & 0.21/0.36 \\
        EigenTrajectory \cite{bae2023eigentrajectory}      & 0.36/0.53      & 0.12/0.19    & 0.24/0.43      & 0.19/0.33       & 0.14/0.24    & 0.21/0.34 \\
        TUTR \cite{shi2023trajectory}      & 0.40/0.61      & \textbf{0.11}/0.18    & 0.23/0.42      & 0.18/0.34       & 0.13/0.25    & 0.21/0.36 \\
        \hline
        PPT (Ours)        & 0.36/\textbf{0.51}      & \textbf{0.11}/\textbf{0.15}    & 0.22/0.40      & \textbf{0.17}/0.30       & \textbf{0.12}/\textbf{0.21}    & \textbf{0.20}/\textbf{0.31} \\
        \bottomrule[2pt]
    \end{tabular}}
    \label{tab:ETH_UCY}
\end{table*}

\begin{table*}[!ht] 
\renewcommand{\arraystretch}{1.1}
    \centering
    \caption{Comparisons with the current state-of-the-art methods on the GCS dataset in minADE$_{20}$ / minFDE$_{20}$ (pixels) metric. Text in \textbf{bold} denotes the best results. Our PPT method significantly outperforms other approaches.}
    \resizebox{\textwidth}{!}{
    \begin{tabular}{c|c|c|c|c|c|c|c|c|c}
        \toprule[2pt]
        Method & \makecell[c]{Social\\-GAN \cite{gupta2018social}} & \makecell[c]{PECNet\\ \cite{mangalam2020not}} & \makecell[c]{Social-\\STGCNN\\ \cite{mohamed2020social}} & \makecell[c]{SGCN\\ \cite{shi2021sgcn}} & \makecell[c]{Agent-\\Former \cite{yuan2021agentformer}} & \makecell[c]{NPSN\\ \cite{bae2022non}} & \makecell[c]{GP-Graph\\ \cite{bae2023graphtern}} & \makecell[c]{Eigen-\\Trajectory\\ \cite{bae2023eigentrajectory}} & PPT (Ours) \\
        \hline
        ADE ↓  & 15.85   & 17.08   & 14.72  & 11.18    & 10.18   & 7.66    & 7.8   & 7.42      & \textbf{6.20}             \\
        FDE ↓  & 32.57   & 29.30   & 23.87  & 20.65   & 16.91  & 13.41   & 13.7  & 12.49       & \textbf{9.34}             \\
        \bottomrule[2pt]
    \end{tabular}}
    \label{tab:GCS}
\end{table*}

\subsection{Comparison with State-of-the-Art Methods}

We quantitatively compare our proposed Progressive Pretext Task learning (PPT) framework with a wide range of current approaches on various datasets. The results show that our framework consistently achieves state-of-the-art (SOTA) performance, particularly surpassing the existing state-of-the-art methods by a significant margin, more than 0.58/1.01 and 1.22/3.15 in ADE/FDE metric on the SDD and GCS datasets, respectively.

On the Stanford Drone Dataset (SDD), we compare our framework with 8 existing methods, which is demonstrated in Table \ref{tab:SDD}. As can be seen, our approach considerably improves the system performance, which reduces the ADE metric from 7.61 to 7.03 and reduces the FDE metric from 11.66 to 10.65 as compared to the current state-of-the-art methods. This illustrates the effectiveness of employing the three-stage progressive pretext tasks for learning short-term dynamics and long-term dependencies, incrementally equipping the model with the ability to predict the entire future trajectory.

\begin{figure}[!t]
  \begin{minipage}[!t]{0.45\linewidth}
    \centering
    \includegraphics[width=\linewidth]{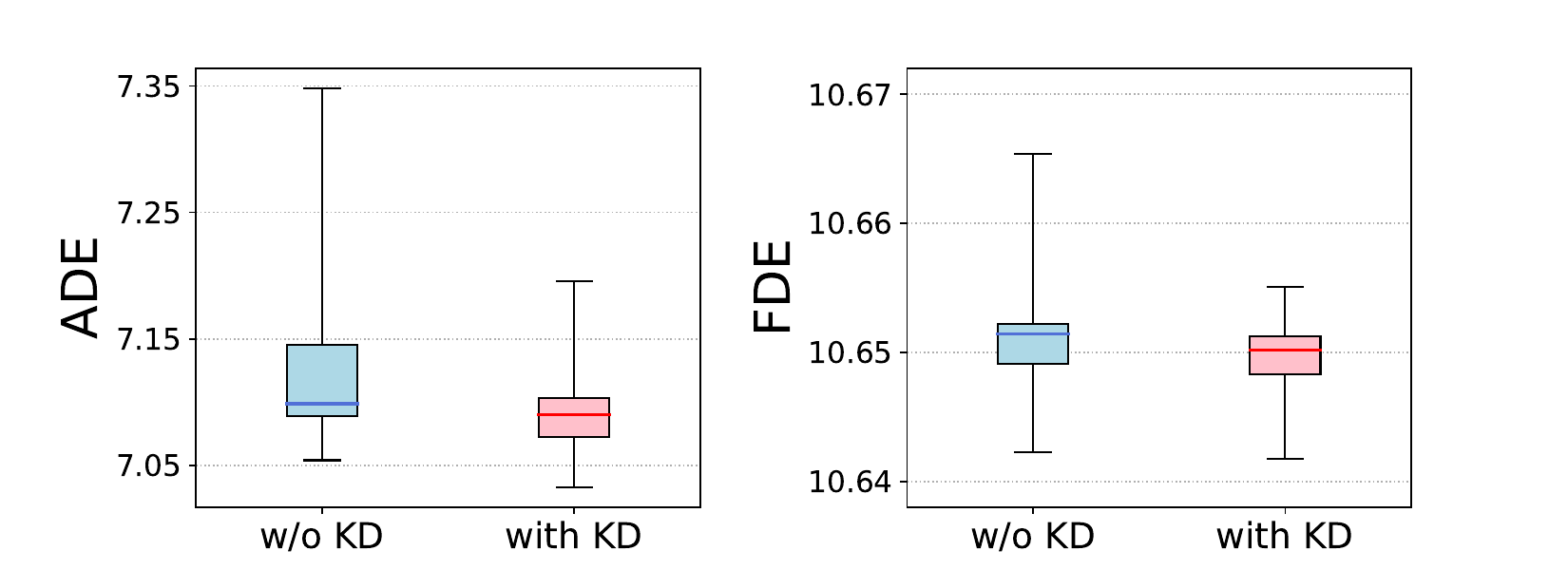}
    \caption{Analysis on the utility of the cross-task knowledge distillation (KD). With cross-task knowledge distillation, the model can produce accurate future trajectories more consistently.}
    \label{fig:KD}
  \end{minipage}%
  \hfill
  \begin{minipage}[!t]{0.5\linewidth}
    \centering
    \renewcommand{\arraystretch}{1.1}
    \centering
    \vspace{-5mm}
    \captionof{table}{The analysis of the pretext tasks on the SDD dataset. We investigate the performance of our framework with neither Task-\uppercase\expandafter{\romannumeral1} nor Task-\uppercase\expandafter{\romannumeral2}, with only Task-\uppercase\expandafter{\romannumeral2}, and with both Task-\uppercase\expandafter{\romannumeral1} and Task-\uppercase\expandafter{\romannumeral2}.}
    \vspace{2mm}
    \resizebox{0.93\textwidth}{!}{
    \begin{tabular}{c c c|c|c}
    \hline
    \hline
        Task-\uppercase\expandafter{\romannumeral1}  & Task-\uppercase\expandafter{\romannumeral2}   & Task-\uppercase\expandafter{\romannumeral3}   & ADE ↓ & FDE ↓\\
        \hline
        \ding{55}  & \ding{55}    & \checkmark   & 10.40   & 18.64      \\
        \ding{55}  & \checkmark   & \checkmark   & 7.71    & 11.42     \\
        \checkmark & \checkmark   & \checkmark   & \textbf{7.03}  & \textbf{10.65}    \\
        
    \hline
    \hline
    \end{tabular}
    }
    \label{tab:PT}
  \end{minipage}
\end{figure}

While on the ETH/UCY dataset, we compare our method with 10 existing approaches. As shown in Table \ref{tab:ETH_UCY}, our progressive pretext task learning framework achieves the best prediction performance again, reducing the average FDE performance from 0.33 to 0.31 and the average ADE performance from 0.21 to 0.20 respectively, compared to the current state-of-the-art methods.

On the Grand Center Station dataset (GCS), we compare the proposed framework with 8 recent approaches. The results in Table \ref{tab:GCS} demonstrate that our progressive pretext task learning framework significantly outperforms the current state-of-the-art method, EigenTrajectory \cite{bae2023eigentrajectory}, by 16.4\% and 25.2\% in ADE and FDE metrics respectively, further verifying the superiority of our PPT framework for the future trajectory prediction.

\subsection{Ablation Studies}
\label{subsec:ablation}
We further conduct ablation studies on the SDD dataset to comprehensively analyze and study the influence of different components in our PPT framework, including the pretext tasks, the cross-task knowledge distillation, and the diversity loss leveraged in Task-\uppercase\expandafter{\romannumeral2}.

\textbf{Effect of the Progressive Pretext Tasks.} 
In Table \ref{tab:PT}, we evaluate the influence of the employed progressive pretext tasks, i.e., Task-\uppercase\expandafter{\romannumeral1} and Task-\uppercase\expandafter{\romannumeral2}, on the system performance. Specifically, we first train the model with all three prediction tasks and then sequentially remove Task-\uppercase\expandafter{\romannumeral1} and Task-\uppercase\expandafter{\romannumeral2} for comparisons. 
As can be observed, both the pretext tasks contribute positively to improving the system performance. Additionally, our experiment shows that with Task-\uppercase\expandafter{\romannumeral1}, the destination prediction performance in Task-\uppercase\expandafter{\romannumeral2} improves from 11.58 to 10.70 in FDE metric. We attribute these to the fact that: i) with Task-\uppercase\expandafter{\romannumeral1}, the model can effectively capture the short-term dynamics inherent in pedestrian trajectory modeling, which contributes a lot to the prediction accuracy. ii) The utilization of Task-\uppercase\expandafter{\romannumeral2} provides the guiding reference and the knowledge of long-term dependencies for the ultimate trajectory sequence prediction, thus significantly improving the prediction performance in both FDE and ADE metrics.

\textbf{Analysis on the Cross-Task Knowledge Distillation.} To examine the effectiveness of employing the cross-task knowledge distillation (KD), we compare the prediction performance of the models trained with and without KD. Over 20 independent runs (with different random seeds) are conducted for each model, and the experimental results are reported in the form of a boxplot in Figure \ref{fig:KD}. As shown, the model trained with KD achieves better prediction performance with smaller variance in both ADE and FDE metrics, suggesting the effectiveness of cross-task knowledge distillation in achieving prediction stability.

\begin{figure}[!t]
  \begin{minipage}[!t]{0.5\linewidth}
    \centering
    \includegraphics[width=\linewidth]{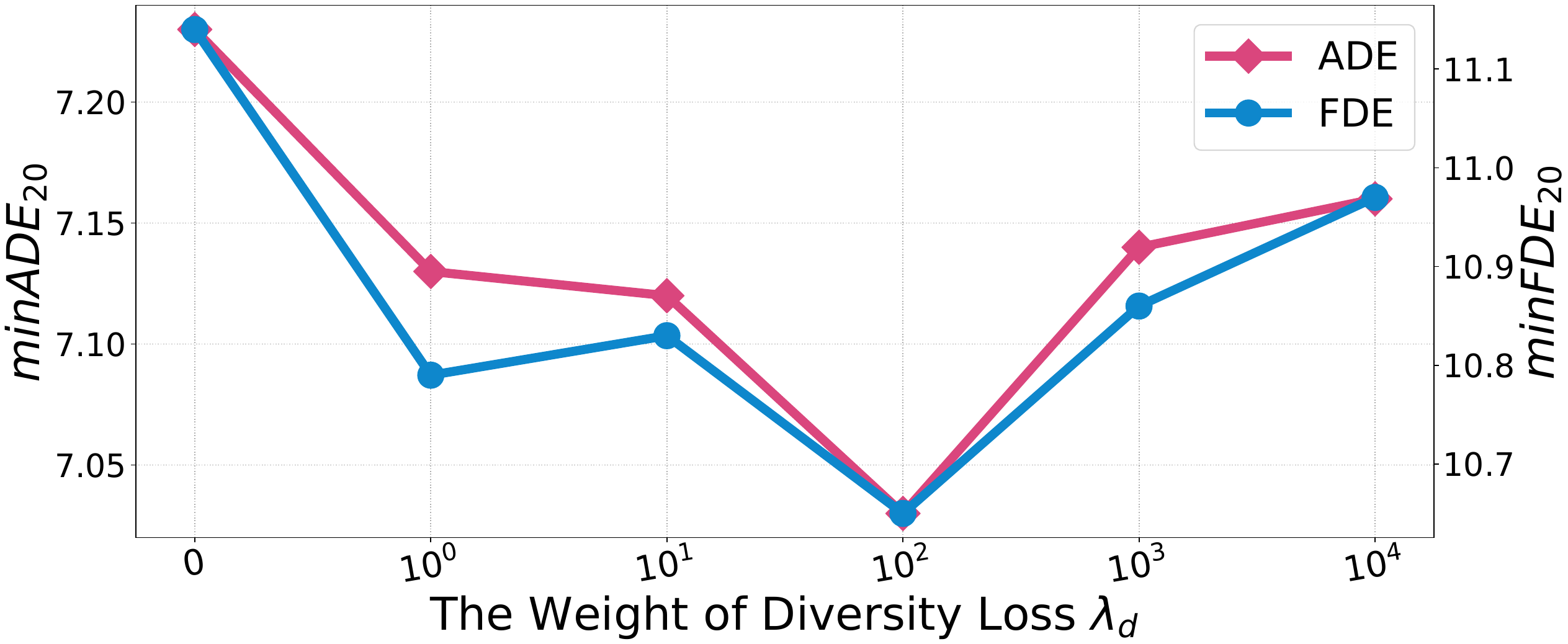}
    \caption{ADE/FDE as a function of the weight $\lambda_d$ in Equation \ref{eq:Des}. $\lambda_d$=100 provides the best performance.}
    \label{fig:DLoss_line}
  \end{minipage}
  \hfill
  \begin{minipage}[!t]{0.45\linewidth}
    \centering
    \includegraphics[width=\linewidth]{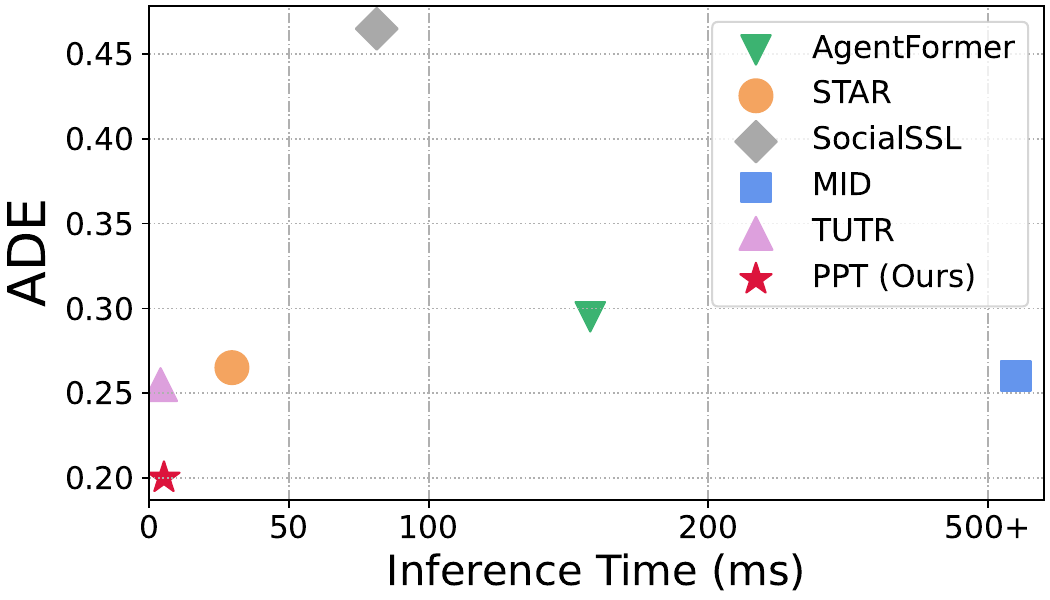}
    \caption{Inference speed and prediction accuracy of Transformer-based models.}
    \label{fig:Inf_time}
  \end{minipage}%
\end{figure}

\textbf{Weight for the Diversity Loss.} 
Figure \ref{fig:DLoss_line} illustrates the influence of different weight $\lambda_d$ (in Equation \ref{eq:Des}) on the prediction performance. As can be observed, the system achieves the best performance when weight $\lambda_d$=100. Either too small or too large $\lambda_d$ leads a performance degradation.
This is because i) when $\lambda_d$ is too small, the model tends to miss the intention modality of the pedestrians, leading to inefficient diversity and worse prediction performance; and ii) when $\lambda_d$ is too large, the impact of the diversity loss gradually dominates the training process. Therefore, the model tends to sacrifice precision for minimizing the diversity loss, resulting in a decrease in prediction accuracy.

\textbf{Efficiency of PPT.} To verify the efficiency of our PPT, we first conduct a comparative analysis of its inference time against five existing Transformer-based approaches. As shown in Figure \ref{fig:Inf_time}, 1) Leveraging the proposed learnable prompt embedding for efficient parallel generation, our predictor achieves an inference speed that significantly surpasses all autoregressive prediction models and remains comparable to the one-step prediction model TUTR \cite{shi2023trajectory} (5.28ms vs. 4.06ms). Also, 2) Trained through our Progressive Pretext Task learning framework, our predictor consistently outperforms all the existing Transformer-based methods in performance. Furthermore, we note that pretraining in earlier stages accelerates convergence in subsequent stages, thus making our PPT framework highly efficient in training time, e.g., 4.7 hours on the SDD dataset. All these results validate the high efficiency and strong effectiveness of our proposed model.

\subsection{Qualitative Results}
 
In this subsection, we provide some visualization results to verify our PPT framework and compare it with the current state-of-the-art approaches qualitatively.

\begin{figure}[!t]
\centering
\includegraphics[width=0.81\linewidth]{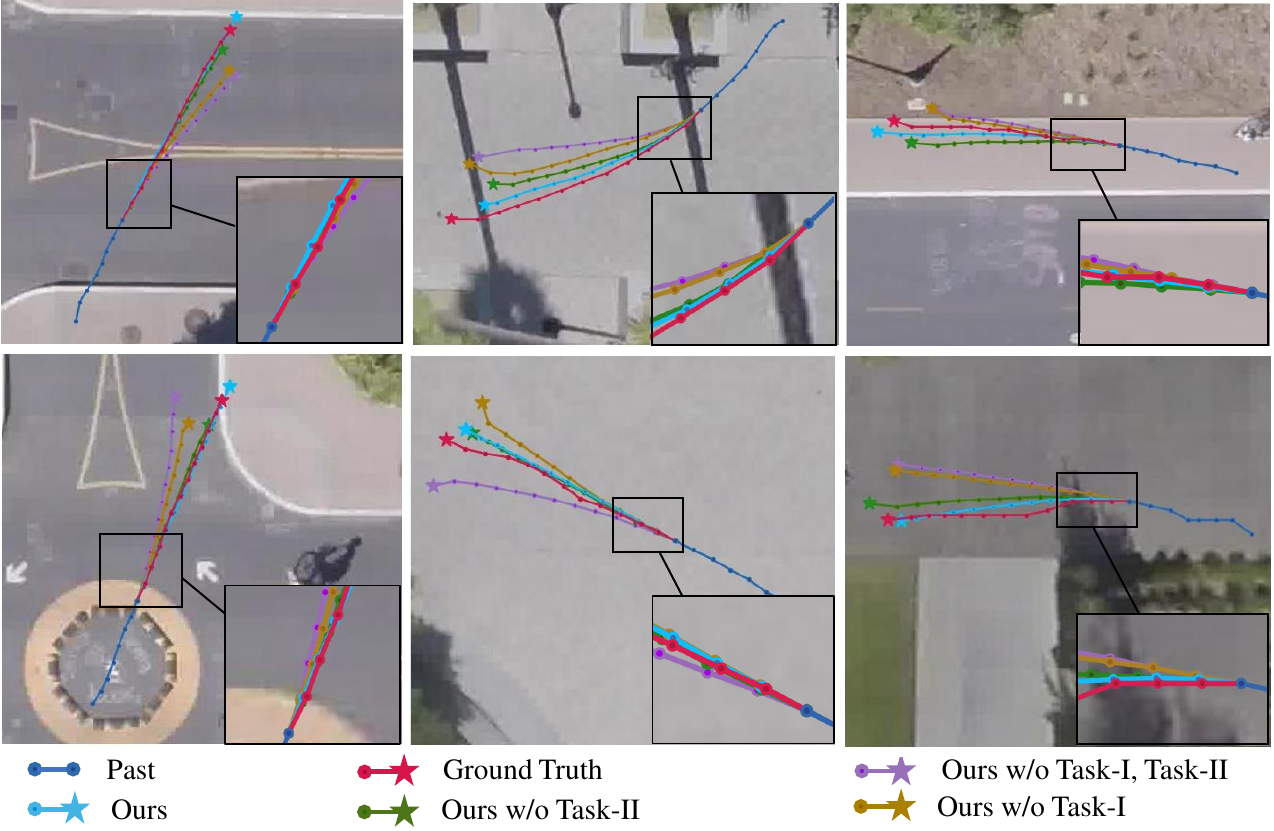}
\caption{Qualitative analysis on the pretext tasks. Our model trained with both Task-\uppercase\expandafter{\romannumeral1} and Task-\uppercase\expandafter{\romannumeral2} can produce more accurate and temporally acceptable trajectories.}
\label{fig:Vis_PPT}
\end{figure}

\textbf{Analysis on the Progressive Pretext Tasks.} We carefully examine the future trajectories predicted by our framework trained with or without pretext Task-\uppercase\expandafter{\romannumeral1} and Task-\uppercase\expandafter{\romannumeral2}. As shown in Figure \ref{fig:Vis_PPT}, on one hand, when pretrained with pretext Task-I, the model can produce more accurate near-future trajectories, validating the effectiveness of using pretext Task-I in capturing the short-term dynamics. On the other hand, better long-term prediction performance is achieved by using pretext Task-II, which suggests that the utilization of Task-II contributes a lot to capturing the long-term dependencies. Furthermore, with both pretext Task-\uppercase\expandafter{\romannumeral1} and Task-\uppercase\expandafter{\romannumeral2}, our framework can visually generate more accurate and more temporally acceptable future trajectories, demonstrating the effectiveness of each progressive pretext task in our PPT framework.

\begin{figure*}[!t]
\centering
\includegraphics[width=\textwidth]{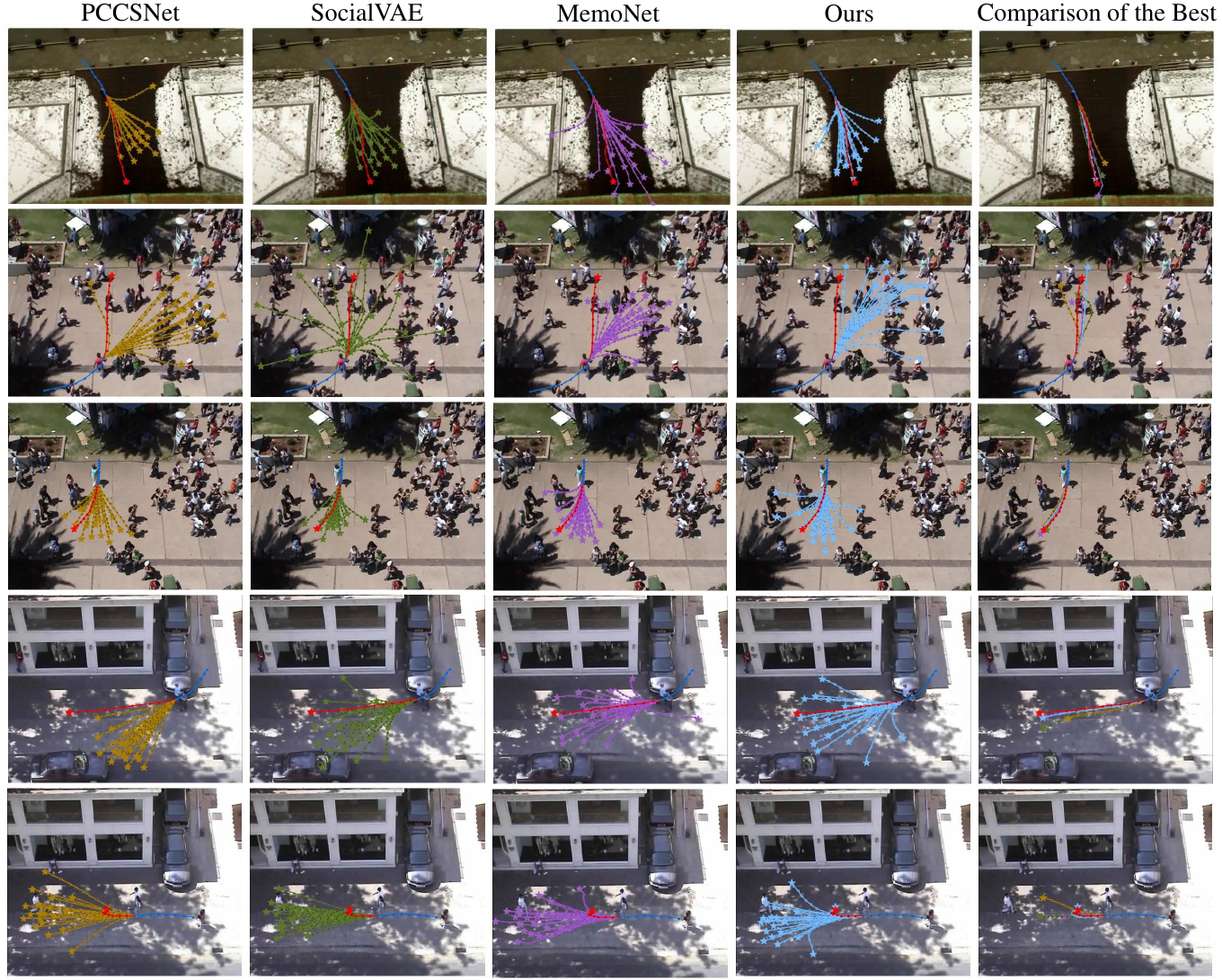}
\caption{The visualization of predicted trajectories on the ETH/UCY Dataset. Each row shows a sample in different scenes. The first four columns illustrate the 20 trajectories predicted by PCCSNet \cite{sun2021three}, SocialVAE \cite{xu2022socialvae}, MemoNet \cite{xu2022remember}, and our PPT framework. The last column demonstrates the best of 20 predictions produced by these approaches. Trajectories in \textcolor{Red}{red} represent ground truth (GT) future trajectories.}
\label{fig:vis_comp}
\end{figure*}

\textbf{Comparison with others.}
Figure \ref{fig:vis_comp} visualizes the future trajectories in the scenes of ETH/UCY datasets predicted by four different approaches, including PCCSNet \cite{sun2021three}, SocialVAE \cite{xu2022socialvae}, MemoNet \cite{xu2022remember} and our PPT framework. The last column illustrates the best of 20 predictions generated by these approaches. The results indicate that among all methods, the future trajectories predicted by our PPT best fit the ground truth future trajectories, validating the superiority of our proposed framework visually. In a more detailed analysis, the first four columns demonstrate the 20 future trajectories predicted by these four methods correspondingly. We observe that compared to other methods, our PPT framework exhibits greater variability in destination predictions while simultaneously maintaining prediction accuracy, thereby generating more accurate and diverse future trajectories. Furthermore, when provided with a destination, a pedestrian typically exhibits a relatively uniform pace toward this destination. As shown, our method can produce future trajectories that are more in line with this motion pattern, compared to MemoNet~\cite{xu2022remember}. This verifies the effectiveness of learning and understanding the temporal dynamics, particularly short-term dynamics and long-term dependencies, in our PPT framework for human trajectory modeling.

\section{Conclusion}
\label{sec:conclu}

In this paper, we present a novel progressive pretext task learning (PPT) framework to formulate pedestrian trajectory prediction, addressing the limitations of previous works by effectively capturing short-term dynamics and long-term dependencies within trajectories. The PPT consists of three stages of progressive training tasks to enhance the model's capacity. Task-\uppercase\expandafter{\romannumeral1} aims to equip the model with the basic ability to comprehend short-term dynamics inherent in the trajectories. Task-\uppercase\expandafter{\romannumeral2} intends to enhance the model to capture long-term dependencies. In Task-\uppercase\expandafter{\romannumeral3}, we finetune the model for the entire future trajectory prediction, exploiting the previously acquired knowledge. A cross-task knowledge distillation is introduced to preserve the knowledge from previous pretext tasks. Further, we design a Transformer-based predictor to complement our framework, which achieves great efficiency with a two-step inference. Extensive experiments are conducted to demonstrate the superiority of our elaborately devised framework.

\noindent \textbf{Acknowledgements.} This work was supported partially by the NSFC (U21A20-471, U22A2095, 62076260, 61772570), Guangdong Natural Science Funds Project (2023B1515040025), Guangdong NSF for Distinguished Young Scholar (2022B15-15020009), Guangdong Provincial Key Laboratory of Information Security Technology (2023B1212060026), and Guangzhou Science and Technology Plan Project (202201011134).


%
%
\bibliographystyle{splncs04}
\bibliography{main}
\end{document}